\documentclass[graybox]{svmult}

\usepackage{type1cm}        %
\usepackage{makeidx}         %
\usepackage{graphicx}        %
\usepackage{multicol}        %
\usepackage[bottom]{footmisc}%

\usepackage{newtxtext}       %
\usepackage[varvw]{newtxmath}       %

\usepackage{algorithm}
\usepackage{algorithmic}
\usepackage{amsmath}
\usepackage{amsfonts}
\usepackage{enumitem}

\DeclareMathOperator*{\argmax}{arg\,max}

\makeindex             %

\begin{document}

\title*{Neural Bandits for Data Mining: Searching for Dangerous Polypharmacy}
\author{Alexandre Larouche, %
    Audrey Durand, %
    Richard Khoury and %
    Caroline Sirois%
}
\institute{Alexandre Larouche \at Universit\'e Laval,  2325 Rue de l'Universit\'e, Qu\'ebec, QC, G1V 0A6,
    \\ \email{alexandre.larouche.7@ulaval.ca}
    \and Audrey Durand \at Universit\'e Laval, 2325 Rue de l'Universit\'e, Qu\'ebec, QC, G1V 0A6
    \\ \email{audrey.durand@ift.ulaval.ca}
    \and Richard Khoury \at Universit\'e Laval, 2325 Rue de l'Universit\'e, Qu\'ebec, QC, G1V 0A6
    \\ \email{richard.khoury@ift.ulaval.ca}
    \and Caroline Sirois \at Centre d'excellence sur le vieillissement de Qu\'ebec 1050 Chemin Ste-Foy, Qu\'ebec, QC, G1S 4L8
    \\ \email{caroline.sirois@pha.ulaval.ca}}
\maketitle

\abstract{
    Polypharmacy, most often defined as the simultaneous consumption of five or more drugs at once, is a prevalent phenomenon in the older population. Some of these polypharmacies, deemed inappropriate, may be associated with adverse health outcomes such as death or hospitalization. Considering the combinatorial nature of the problem as well as the size of claims database and the cost to compute an exact association measure for a given drug combination, it is impossible to investigate every possible combination of drugs. Therefore, we propose to optimize the search for potentially inappropriate polypharmacies (PIPs). To this end, we propose the OptimNeuralTS strategy, based on Neural Thompson Sampling and differential evolution, to efficiently mine claims datasets and build a predictive model of the association between drug combinations and health outcomes. We benchmark our method using two datasets generated by an internally developed simulator of polypharmacy data containing 500 drugs and 100 000 distinct combinations. Empirically, our method can detect up to 72\% of PIPs while maintaining an average precision score of 99\% using 30 000 time steps.
}

\section{Introduction}
Polypharmacy is most often defined as the simultaneous consumption of five or more drugs at once by a patient~\cite{masnoon2017polypharmacy} and is a prevalent phenomenon in the older population. In the USA, 65.1\% of older adults experience polypharmacy, with most of them using more than 5 medications at once ~\cite{young2021polypharmacy}. In Canada, older adults in long term care facilities use on average 9.9 drug classes, while older adults living outside of these facilities use on average 6.7 drug classes~\cite{canadian2018drug}.
Some polypharmacies can be dangerous, in the sense that they can lead to to negative health outcomes, like death or hospitalization. Fortunately, screening tools exists to avoid prescription of potentially inappropriate drugs (e.g., opioids and benzodiazepines~\cite{20192019american}). These tools essentially consist of finite lists of individual drugs and drug pairs that have been identified as dangerous by experts during pharmaco-epidemiological studies as well as from experience. By definition, this means that potentially dangerous combination resulting from more than two drugs interacting together cannot be identified using the screening tools, in addition to other currently unknown dangerous drug combinations. In order to prevent the prescription of potentially harmful polypharmarcies, it would be important to expand the screening tools until they ideally contain all potentially dangerous combinations. Unfortunately, given all the possible drug combinations as well as their varying effects depending on different patient characteristics, pharmaco-epidemiologists cannot investigate all of them.

The goal of this work is therefore to build a predictive model able to identify drug combinations at risk of being harmful, so that they can be investigated further. We propose to achieve this by leveraging neural networks to predict an association measure to a health outcome given any input describing an arbitrary number of drugs.
In practice, such model would be trained using historical data on drugs prescribed to patients, their clinical and sociodemographic characteristics, and their health outcomes. These datasets are typically very large, which makes the association measure expensive to compute, and highly unbalanced.

We therefore tackle the general problem of efficiently mining historical data to train a generalizable and useful model. To achieve this, we formulate the problem under the neural bandit setting so that it can be addressed with the Neural Thompson Sampling (NeuralTS)~\cite{zhang2021neural} strategy. However, using this strategy on a very large action space (such as the one considered here) also raises challenges, which we address by combining NeuralTS with differential evolution (DE)~\cite{storn1997differential}.
The proposed OptimNeuralTS approach finally results in an ensemble predictor made of the evolving sequence of models trained on all the intermediate data subsets.
We evaluate the potential of OptimNeuralTS in simulated experiments. Our results show that our approach can be used to iteratively build an information-rich dataset that can in turn be used for training a predictive model, resulting in an ensemble model capable of extracting new potentially inappropriate polypharmacies (PIPs).
We finally provide an overview of related work in machine learning (in general) applied to polypharmacy discovery and bandit strategies applied for data mining.
We highlight two contributions:
\begin{enumerate}
    \item Tackling the problem of efficient creation of information-rich datasets under the contextual bandit setting.
    \item Introducing the OptimNeuralTS approach to learn predictive models by mining relevant data from very large unbalanced datasets.
\end{enumerate}

\section{Problem formulation}

Let $\mathcal D$ denote a historical dataset containing information about drug combinations and health outcomes. Table~\ref{tab:hist_data} shows a simple example of such a historical dataset, where each line corresponds to a drug combination identified in a binary format, (i.e. 0/1 indicate whether a drug was taken or not by an individual) along with a binary variable indicating whether the individual developed (or not) a given health outcome while consuming the drug combination.
\begin{table}[b]
    \caption{Simple example of historical dataset.}
    \label{tab:hist_data}
    \centering
    \begin{tabular}{l @{\hspace{8pt}} ccccc}
        \svhline
        ID       & Drug 1   & Drug 2   & ...      & Drug N   & Outcome  \\
        \hline
        \hline
        1        & 1        & 0        & ...      & 1        & 0        \\
        1        & 1        & 0        & ...      & 1        & 1        \\
        2        & 0        & 1        & ...      & 1        & 1        \\
        $\vdots$ & $\vdots$ & $\vdots$ & $\ddots$ & $\vdots$ & $\vdots$ \\
        \hline
    \end{tabular}

\end{table}
In the targeted application, the measure of association is called the relative risk (RR) and is described as the risks for a given health outcome in the exposed population over the risks in the unexposed population~\cite{sistrom2004proportions}. An exposed patient is a patient which takes a given drug combination. The exposed population is simply the portion of the population which consumes the combination of drugs. Mathematically, given Table~\ref{tab:contingency}, the RR is defined as:
$$
    \operatorname{RR} = \frac{a(c + d)}{c(a + b)}.
    \label{eqn:RR}
$$
\begin{table}[b]
    \caption{Example of contingency table. An exposed patient is a patient taking a given combination of drugs. A health outcome is computed as a yes if the health outcome occurred during the duration of the prescription of the drug combination (1 in the corresponding line in Table~\ref{tab:hist_data})}
    \label{tab:contingency}
    \centering
    \begin{tabular}[b]{l @{\hspace{8pt}} cc}
        \hline
                    & \multicolumn{2}{c}{Health Outcome}      \\
                    & Yes                                & No \\
        \hline
        \hline
        Exposed     & a                                  & b  \\
        Not Exposed & c                                  & d  \\
        \hline
    \end{tabular}

\end{table}

RR has the advantage of having an implicit threshold: if RR $> 1$ then a drug combination is associated with a given health outcome, therefore it is potentially harmful, while RR $< 1$ implies that a drug combination protects against a given health outcome, therefore it is safe. RR $= 1$ means that a drug combination is neither protective against nor associated with a given health outcome.

In practice, historical datasets are extracted from medico-administrative databases. Therefore, medication claims data typically contain hundreds of columns related to drug usage and millions of rows enumerating all possible simultaneous drug combinations taken by all patients in a large population, but no precomputed association measure. Unfortunately, computing the association measure for a given drug combination is computationally expensive since it requires enumerating every row in the dataset. Therefore, as the size of the data grows, it becomes harder to compute. This raises a challenge when aiming to train a model capable of predicting an association measure to the considered health outcome for any drug combination provided as input because it is not computationally realistic to compute the RR for the whole dataset $\mathcal D$.
The training of a predictive model must therefore be performed on a subset of at most $T$ samples from $\mathcal D$, where $T \ll |\mathcal D|$. This opens the question of how to sample such subsets from $\mathcal D$.

\begin{figure}[t]
    \sidecaption[t]
    \centering
    \includegraphics[width=0.5\textwidth]{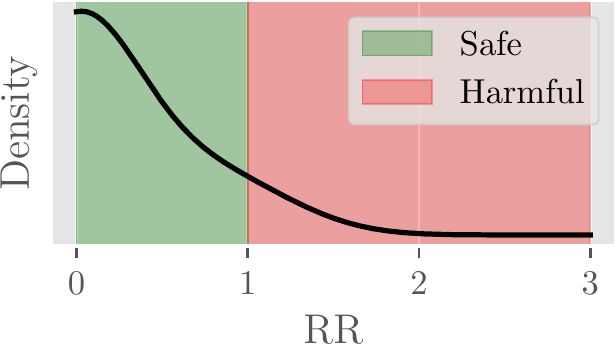}
    \caption{Example of a typical distribution the estimated RR for historical data. A drug combination with a RR $> 1$ is considered harmful while if RR $< 1$, the combination is considered safe.}
    \label{fig:measure_dist}
\end{figure}

In addition, one must note that the historical dataset $\mathcal D$ is highly unbalanced.
Indeed, prescribers usually prescribe drugs which do not interact together in a harmful way, thus most drug combinations have a low measure of association to health outcomes such as hospitalization and death. Figure~\ref{fig:measure_dist} displays a simulated but typical distribution of the estimated association measures observed in the real data.
Therefore, randomly sampling from $\mathcal D$ would yield a training dataset containing mostly combinations of drugs with low measures of association as PIPs associated with adverse health outcomes are rare. However, it is well known that the performance of a predictive model highly depends on the quality of the underlying data~\cite{gudivada2017data}. In other words, if the training dataset contains few to no PIPs associated with adverse health outcomes, the predictive model is very unlikely to learn to identify such PIPs. Therefore we need a strategy capable of sampling a training dataset with higher odds of containing PIPs associated with adverse health outcomes.

\section{Proposed Approach}

We tackle this challenge by formulating the training dataset creation problem as a contextual bandit problem ~\cite{langford2007epoch}, where we leverage the NeuralTS~\cite{zhang2021neural} action selection strategy combined with DE~\cite{storn1997differential} to select drug combinations for which to compute the RR.

\subsection{Neural Contextual Bandits}

A contextual bandit environment is described by a collection of actions $\mathcal{A}$, a feature space $\mathcal{X}$, and an unknown reward function $h: \mathcal X \mapsto \mathbb{R}$, such that each action $a\in\mathcal A$ is associated with a feature vector, or context, $x_a\in\mathcal X$. At each time step $t=1, 2, \dots, T$ of the contextual bandit game, the player (agent interacting with the environment) is presented with a subset of actions $\mathcal{A}_t \subset \mathcal{A}$. It then selects an action $a_t\in\mathcal{A}_t$ to play and observes a noisy reward $r_t = h(x_{a_t}) + \xi_t$, where $\xi_t$ is a $\sigma$-sub-Gaussian noise (e.g., $\xi_t \sim \mathcal{N}(0, \sigma)$). The goal of an agent playing this game is to maximize the cumulative reward, defined as:
\begin{align}
    \sum\limits_{t = 1}^{T} r_t
    \label{eqn:obj_bandit}
\end{align}
The obvious solution is to simply play the optimal action $a_t^*$ at every round, which has the highest expected reward for this round. That is $a^*_t = \argmax\limits_{a \in \mathcal{A}_t} h(x_a)$. However, function $h$ is not known a priori, therefore, the agent must learn by trial-and-error in order to improve its behavior over time.

In the training dataset construction problem from historical data, $\mathcal{A}$ corresponds to the set of all possible drug combinations, $\mathcal{A}_t$ corresponds to the set of drug combinations that are available to explore at time $t$, the features $x_a$ correspond to a multi-hot representation of drug combination $a$, and the reward function $h$ corresponds to the measure of association between a drug combination and a given health outcome, i.e., the RR. At each time step $t$, the agent selects a drug combination $a_t$ for which to compute the RR.
Since computing the RR on the historical dataset is computationally expensive, it is instead computed on a subset of the data, hence the noisy reward $r_t$.

Neural bandit strategies, such as  NeuralUCB~\cite{zhou2020neural} and NeuralTS~\cite{zhang2021neural}, rely on a neural network $f(\cdot;\theta) : \mathcal X \mapsto \mathbb{R}$ to model the reward function $h$ in order to predict the expected reward given any feature $x \in \mathcal{X}$. More importantly, these approaches can estimate the confidence interval around the prediction of the neural network to guide the exploration. They achieve this by using the gradient on the activation, $g(\cdot; \theta) : \mathcal X \mapsto \mathbb{R}^{|\theta|}$.

\subsubsection{NeuralTS}

NeuralTS uses the gradient to estimate the distribution of reward for an action. Indeed, at step $t$, the parameters of a normal prior $\mathcal N(f_t, s_t)$ are estimated as follows:
\begin{align}
    f_t(\cdot) & = f(\cdot; \theta_{t})
    \qquad \text{and}
    \label{eqn:mean}                                                                               \\
    s_t(\cdot) & = \sqrt{\lambda g(\cdot; \theta_{t})^{\top} U_{t}^{-1} g(\cdot; \theta_{t})} / m,
    \label{eqn:std}
\end{align}
where $U_t = \lambda \mathbf{I}_m + \sum\limits_{i = 1}^t g(x_{a_i}, \theta_{i-1})g(x_{a_i}, \theta_{i-1})^\top$ is a design matrix containing the gradient computed for the inputs (selected actions) up to time $t$, $\lambda$ is a regularization parameter and $m$ is the number of parameters in the neural network.
NeuralTS selects the action $a_t$ to play by sampling a value from $\mathcal{N}(f_t(x_{a}), s_t(x_{a})), \forall a \in \mathcal{A}_t$ and picking the action with the highest value. After an action $a_t$ is played, the associated context $x_{a_t}$ as well as the observed reward $r_t$ are added to the training dataset $\mathcal D_t = \mathcal D_{t-1} \cup \{ (x_{a_t}, r_t) \} $ used for computing the new network parameters $\theta_t$, before moving on to the next time step.

Since the bandit strategy will seek to play actions which yield high rewards, this should lead to a dataset $\mathcal D_T$ containing a reasonable amount of drug combinations with high measures of association to a given health outcome. We therefore hypothesize that such a dataset will make it possible to train a predictive model capable of identifying PIPs with high precision. However, for the bandit strategy to be able to recommend actions with high RR, such actions need to be contained in the set of available actions $\mathcal{A}_t$.

\subsection{Generating relevant available action sets}

From the neural contextual bandit problem formulation, action $a_t$ is selected from the subset $\mathcal{A}_t \subset \mathcal A$ containing all available actions at time $t$. This is due to the fact that the bandit strategy must consider each action $a\in\mathcal A_t$ in order to recommend $a_t$, and that the complete action set $\mathcal A$ is typically too large to be entirely considered at every time step. This is also the case in the considered application
due to the combinatorial nature of polypharmacy.
For the same reason that the predictive model training dataset cannot be sampled at random from $\mathcal D$, we cannot generate $\mathcal{A}_t$ by randomly sampling from $\mathcal{A}$ due to the highly skewed distribution of RRs. We must therefore generate subsets $\mathcal A_t$ such that the presence of potentially harmful drug combinations is favored.

To achieve this, we propose to generate subsets of available actions $\mathcal{A}_t$ using differential evolution (DE)~\cite{storn1997differential}, an evolutionary optimization algorithm which does not rely on a gradient signal to converge to a solution. The general principle behind DE is to maintain a population and mutate its members, which are feature vectors, according to a strategy. Here, we consider the best/1/bin strategy~\cite{storn1997differential} described in Algorithm~\ref{alg:de}, where the objective function $q: \mathcal{X} \mapsto \mathbb{R}$ corresponds to an action value function sampled from a neural network. The DE optimization process is therefore conducted \textit{on} function $q$.
\begin{algorithm}[tb]
    \caption{DE best/1/bin}
    \label{alg:de}
    \textbf{Input}: Population size $N$, crossover rate $C$, differential weight $F$, number of optimization steps $S$, objective function $q(\cdot)$ to maximize\\
    \textbf{Output}: Best member $b_{\star}$ in the final step\\
    \begin{algorithmic}[1] %
        \STATE Initialize population $\mathcal{W}$ with $N$ feature vectors sampled from the domain $\mathcal{X}$.
        \FOR{$s\leftarrow1...S$}
        \STATE Let $b \leftarrow \argmax\limits_{w_i \in \mathcal{W}} q(w_i)$
        \FOR{$w_i \in \mathcal{W}$}
        \STATE Sample $v \sim \mathcal{U}(0, 1)$
        \STATE Randomly select indices $l$, $r_1$ and $r_2$ in $[N] - [i]$
        \STATE Generate a new feature vector: $$m_i \leftarrow b + F (w_{r_1} - w_{r_2})$$

        \STATE Generate a mutated feature vector where components $j$ are computed as follows:
        $$u_{i, j} \leftarrow \begin{cases}
                m_{i, j} \textbf{ if } j=l \textbf{ or }  v \leq C \\
                w_{i, j} \textbf{ otherwise}                       \\
            \end{cases}$$

        \STATE $w_i \leftarrow \begin{cases}
                u_i \textbf{ if } q(u_i) \leq q(w_i) \\
                w_i \textbf{ otherwise}
            \end{cases} $
        \ENDFOR
        \ENDFOR
        \STATE $b_{\star} \leftarrow \argmax\limits_{w_i \in \mathcal{W}} q(w_i)$
        \STATE \textbf{return } $b_{\star}$
    \end{algorithmic}
\end{algorithm}

DE with best/1/bin therefore corresponds to considering $|\mathcal A_t| = N\times S$ available actions at each time step $t$. Parameters $N$ and $S$ are typically chosen such that $N\times S \ll |\mathcal A|$. The best member returned after the $S$ steps of DE corresponds to the action features maximizing $q$, which is a value function given by the neural network model. Therefore the best member would correspond to $a_t$. For example, with NeuralTS, $q(\cdot)$ corresponds to a sample from the distribution $\mathcal N(f_{t-1}(\cdot), s_{t-1}(\cdot))$ at time step $t$ (see Eq.~\ref{eqn:mean}~and~\ref{eqn:std}).
Now, computing $r_t$ on-the-fly on $\mathcal{D}$ requires the selected action to be contained in $\mathcal D$. However, DE (best/1/bin) is not constrained to $\mathcal{D}$, so this condition may not be fulfilled. In order to account for this situation, we propose to select the action $a_t$ as being the 1-nearest-neighbor in $\mathcal{D}$ to the action returned by DE (best/1/bin) with ties broken arbitrarily. This ensures that the association measure for the drug combination can be computed.

\subsection{OptimNeuralTS}

Algorithm~\ref{alg:optimneuralts} describes the resulting OptimNeuralTS.
\begin{algorithm}[tb]
    \caption{OptimNeuralTS}
    \label{alg:optimneuralts}
    \textbf{Input}: Dimension of feature vectors $d$, number of time step $T$, number of warm-up steps $\tau$, regularization term $\lambda$, exploration factor $\nu$, number of training epochs $J$, learning rate $\eta$, DE population size $N$, DE crossover rate $C$, DE differential weight $F$, number of DE steps $S$, historical dataset $\mathcal{D}$\\
    \textbf{Output}: Generated dataset $\mathcal D_T = \{(x_{a_{i}}, r_i) \}_{i=1}^T$ and ensemble neural network models $\{\theta_i\}_{i=1}^T$\\
    \begin{algorithmic}[1] %
        \STATE Initialize neural network parameters $\theta_0$
        \STATE $U_0 \leftarrow \lambda I_{|\theta_0|}$ and $\mathcal D_0 \leftarrow \{\}$

        \FOR{$t\leftarrow 1...\tau$}
        \STATE Randomly play action $a_t \in \mathcal{D}$, observe $r_t$
        \STATE $U_t \leftarrow U_{t-1} + g(x_{a_t}; \theta_0)g(x_{a_t}; \theta_0)^\top$
        \STATE $\mathcal D_t \leftarrow \mathcal D_{t-1} \cup \{ (x_{a_t}, r_t) \}$
        \ENDFOR

        \STATE $\theta_{\tau} \leftarrow \text{Train network with parameters } \eta, J, \mathcal D_\tau, \theta_0$

        \FOR{$t\leftarrow \tau+1...T$}
        \STATE $\hat a_{t} \leftarrow$ DE$\left(N, C, F, S, \mathcal{N}\left(f_{t-1}, \nu s_{t-1} \right)\right)$

        \STATE $U_t \leftarrow U_{t-1} + g(x_{\hat a_t}, \theta_{t-1})g(x_{\hat a_t}, \theta_{t-1})^\top$

        \STATE Set $a_{t}$ as the 1-nearest-neighbor of $\hat a_t$ in $\mathcal{D}$

        \STATE Play $a_{t}$, observe $r_t$
        \STATE $\mathcal D_t \leftarrow \mathcal D_{t-1} \cup \{ (x_{a_t}, r_t) \}$
        \STATE $\theta_t \leftarrow \text{Train network with parameters } \eta, J, \mathcal D_{t}, \theta_{t-1}$
        \ENDFOR

        \STATE \textbf{return } $\mathcal D_T$, $\{\theta_i\}_{i=\tau}^T$
    \end{algorithmic}
\end{algorithm}
The agent warms up by randomly sampling actions during the first $\tau$ steps, observing their rewards and updating the internal parameters $U$ (lines 3-7).
The neural network is then trained for the first time on the random data using a standard gradient descent with the L2 regularization scheme of NeuralTS (line 8), before the agent starts playing actions according to the NeuralTS and DE strategy.
We slightly abuse the notation when calling DE (line 10) to indicate that the objective function $q$ evaluated at features $x$ consists in a normal distribution centered at $f_{t-1}(x)$ with standard deviation $s_{t-1}(x)$ (see Eq.~\ref{eqn:mean}~and~\ref{eqn:std}).
The design matrix $U$ of the agent is then updated (line 11),
the agent plays the transformed action $a_t$, observes the reward $r_t$, and updates the dataset $\mathcal{D}_t$ before updating the neural network parameters with the same procedure as previously stated (lines 13-15).
OptimNeuralTS finally returns the dataset generated by the algorithm as well as the ensemble model corresponding to all the intermediate models ($\theta_\tau \dots \theta_T$) encountered along the search (line 17). Indeed, as experiments will show, the subsets of data encountered along the neural contextual bandit game will result in neural network models that are specialized in different relevant regions of PIPs. Combining these models in an ensemble therefore results in a strong predictive model with a good coverage of the input space, which can then be used to predict a RR for any given drug combination. Detecting new PIPs is then only a matter of applying a threshold over the predicted RR's lower confidence bound, which can be computed from $f_t(\cdot)$ and $s_t(\cdot)$.

\subsubsection{Warming-up}

Previous work showed that non-informative priors can impact performance~\cite{mary2014bandits}. To mitigate this, we allow the agent to warm-up by selecting $\tau$ random actions $a \in \mathcal D$ and observe their rewards as a way to initialize its belief about the data. This effectively creates a small randomly sampled dataset composed of the seen contexts and the observed rewards. This is helpful, as the data gathered after this point is dependent on the previously gathered data, therefore breaking the i.i.d. assumption of data in supervised learning. This in turn can lead to a failure in learning as an agent without any representation of the relationship in the data may sample it poorly when playing, leading to a poor representation and so on.

\subsubsection{Transforming recommended actions into playable actions}

As previously mentioned, our problem requires that we transform $\hat a_t$ into $a_t$. However, $U_t$ is updated with $g(x_{\hat a_t}; \theta_{t - 1})$ instead of $g(x_{a_t}; \theta_{t - 1})$ (line 12). Two facts motivate this choice of update. The first is that if the entirety of  $\mathcal{A}$ was present in $\mathcal{D}$, then no transformation would be needed. Indeed, the transformation is only implemented in order to train the neural network on a relationship existing in the historical data and so $\hat a_t$ is the recommended action. Secondly, due to the distribution of RRs on our data, the gradient of $a_t$ often contains very little information. This is due to the fact that the RR is concentrated around the mean, which leads to the last bias term of the neural network being almost the only participating term in the prediction. The gradient vector $g(x_{a_t}; \theta_t)$ is then almost barren, which in turn leads to a design matrix containing little information. In practice, updating $U_t$ with $g(x_{a_t}; \theta_t)$ works, but we have found it leads to much less new PIPs detected much later during the bandit algorithm's training.

\section{Experiments}
\label{sec:experiments}
Evaluating the proposed approach requires a dataset with a ground-truth and a structure similar to the real world data. As no such dataset is readily available, we first develop a simulator to generate synthetic data.

\subsection{Synthetic data generation}

\begin{figure}[t]
    \sidecaption[t]
    \centering
    \includegraphics[width=0.649\columnwidth]{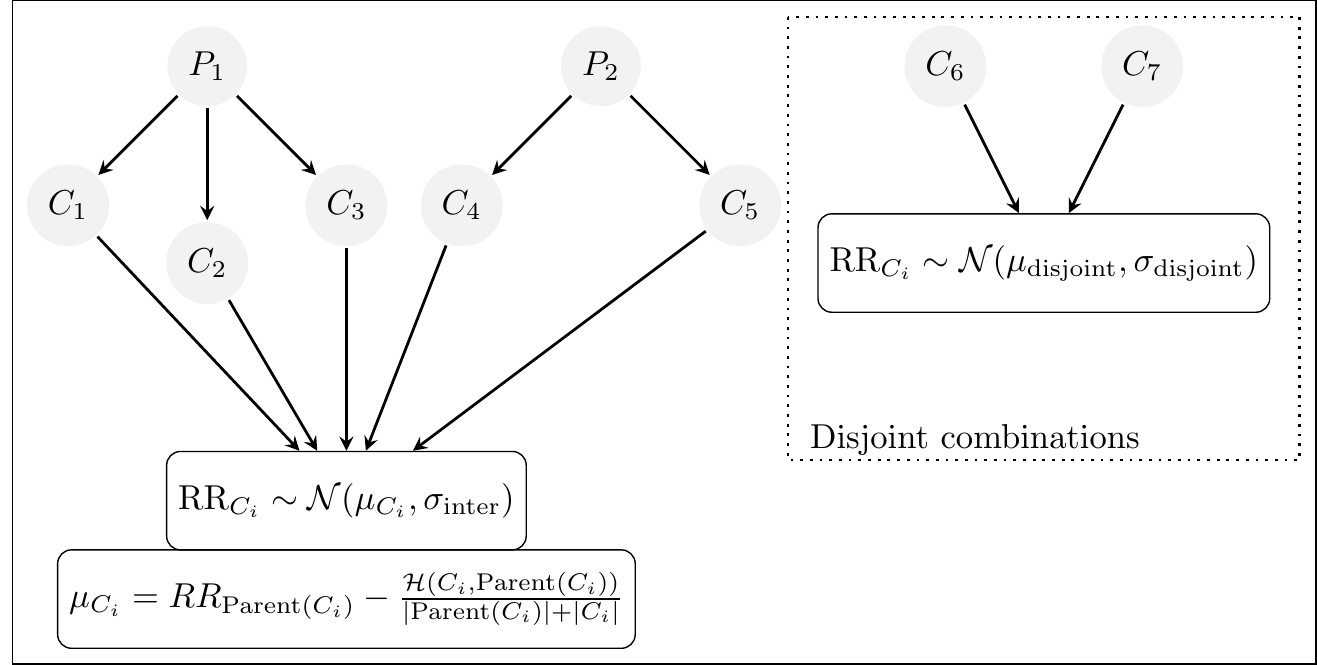}
    \caption{Overview of the simulated assignment of RR to drug combinations. The RR attributed to a combination $C$ is proportional to its similarity to its nearest dangerous pattern $P$, shown as the parent in the resulting tree. $\sigma_\text{inter}, \sigma_\text{disjoint}$ and $\mu_\text{disjoint}$ are user-defined parameters.}
    \label{fig:sim}
\end{figure}

Our main hypothesis guiding data generation is that a drug combination similar to a drug combination with a high RR should have a similarly high RR. Consequently, we generate by sampling from binomial distributions the set $\mathcal P$ of what we call ``dangerous patterns'', which are characterized by a high RR. Likewise, we randomly generate the set of distinct drug combinations $\mathcal C$ without attributing them a RR. The similarity between each drug combination $C\in \mathcal C$ and each dangerous pattern $P\in\mathcal P$ are then computed using the Hamming distance. Two cases can arise from this: 1) either drug combination $C$ has some drug(s) in common with the nearest pattern or 2) it does not. If it does, then the drug combination is said to \textit{intersect with the pattern} and is attributed a RR proportional to its similarity to its nearest pattern $P$. Alternatively, if a combination is disjoint of the nearest pattern, then its RR is sampled from a normal distribution $\mathcal N(\mu_{\text{disjoint}}, \sigma_{\text{disjoint}})$. This procedure results in a dataset with only distinct combinations with a precomputed RR. The Hamming distance, by definition, favors dangerous patterns containing smaller subsets of drugs during the nearest pattern search. As a result, a combination's nearest dangerous pattern is not always the one with the biggest overlap in terms of drugs. This results in a dataset with a RR not necessarily increasing proportionally to the size of the intersection between combinations and patterns, which adds difficulty to the problem. Figure~\ref{fig:sim} gives the overview of the RR generation process.

\subsection{Experimental setup}

We devise one experiment on two datasets generated by the simulator. The goal of the experiment is to detect drug combinations with an RR above a certain threshold. In our work, we consider a threshold of $1.1$ for the RR since we consider a RR between $1.0$ and $1.1$ to be too low to be significant. \textbf{The most important aspect here is not to find every PIP but to detect them with as few false positives as possible.} This is crucial, as in practice these findings need to be further studied by healthcare experts and it is laborious to do so. Furthermore, since we plan on using samples of the real dataset in our practical application, we also add a noise term $\xi_t \sim \mathcal{N}(0, 0.1)$ to the observed RR for a drug combination to simulate sample noise. Therefore, to ensure a low false positive rate, a drug combination $a$ is only classified as potentially harmful if $f_t(x_a) - 3s_t(x_a) > 1.1$, to emulate a pessimistic 99\% lower confidence bound. The choice of $\sigma = 0.1$ for the normal distribution is so the noise does not dominate the reward signal while still resulting in a challenging instance in the datasets described below.

\begin{figure}[t]
    \sidecaption[t]
    \includegraphics[width=0.55\columnwidth]{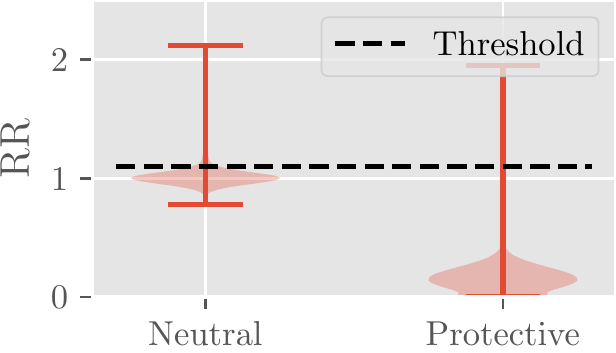}
    \caption{Distribution of RRs for the neutral and protective datasets}
    \label{fig:distributions_datasets}
\end{figure}

We generate two datasets each representing a different hypothesis on the effect of the consumption of drugs: a neutral effect instance,where most RRs are near 1, and a protective effect instance (where most RRs are concentrated near 0). The average RR in the latter is well below the dangerous RR threshold, as would be expected in a real dataset. However, to study the robustness of the proposed approach, we also consider a neutral effect dataset where the distribution of RRs is concentrated around the threshold. This leads to a more challenging instance as the noise $\xi_t$ is more likely to make a safe combination appear potentially harmful.
Figure~\ref{fig:distributions_datasets} shows the distribution of RRs in both datasets. Both datasets contain 100k distinct combinations of 500 possible drugs, with RRs computed from 10 random dangerous patterns that do not appear in the distinct combinations. The two datasets are highly unbalanced, with the neutral and protective datasets respectively containing $2082$ and $7805$ distinct dangerous drug combinations (RR $> 1.1$).

\begin{table}[t]
    \caption{OptimNeuralTS parameter values; left side are specific to the DE component. $^\star\eta$ is reduced when the loss reaches a plateau during training.}
    \label{tab:params}    \centering
    \begin{tabular}{c c @{\hspace{8pt}} c c}
        \multicolumn{2}{c}{DE best/1/bin} & \multicolumn{2}{c}{OptimNeuralTS}                            \\
        Parameter                         & Value                             & Parameter & Value        \\
        \hline
        \hline
        $N$                               & $32$                              & $d$       & $500$        \\
        $C$                               & $0.9$                             & $T$       & $30\text{k}$ \\
        $F$                               & $1$                               & $\tau$    & $10\text{k}$ \\
        $S$                               & $16$                              & $\lambda$ & $1$          \\
                                          &                                   & $\nu$     & $1$          \\
                                          &                                   & $J$       & $100$        \\
                                          &                                   & $\eta$    & $0.01^\star$ \\
        \hline
    \end{tabular}

\end{table}
Table~\ref{tab:params} shows the OptimNeuralTS parameters used in this experiment. The number of time steps $T$ is chosen such that only a small fraction of the entire drug combination space can be investigated during the bandit game. Indeed, in real life applications, millions of drug combinations are typically available. Setting $T$ to a small number compared to the number of possible combinations thus requires OptimNeuralTS to be efficient in its choice of drugs to investigate at every round in order to succeed. The results shown here are for a warm-up duration of $\tau = 10\text{k}$ samples and an exploration factor of $\nu = 1$ taken as the best configuration from a grid search of the space $\tau \in  \{1\text{k}, 10\text{k}, 20\text{k}, 30\text{k}\}$ and $\nu \in \{1, 10\}$\footnote{The total number of configurations tried is thus 7, as $\tau = 30\text{k}$ is the same for any $\nu$}. All the configurations in the grid search succeed in finding variable amounts of PIPs with high precision, except when the warm-up phase is too long (e.g. $\tau = 30\text{k}$), highlighting the need for a bandit strategy. Furthermore, the considered space for the grid search of $\nu$ is never below $1$ to discourage greedy action (i.e. exploiting the already known PIPs). Furthermore, we schedule the learning rate $\eta = 0.01$ to decrease as the loss reaches a plateau during training. This parameter was found by a hyperparameter search guided by OpTuna~\cite{optuna_2019} and Tune~\cite{liaw2018tune}. As for DE, the parameters were selected manually to maximize the mutation of the population at each optimization step while still maintaining a very small population and a quick runtime.

As previously mentioned, by applying a threshold over the lower bounds on neural network predictions, the regression problem of learning a mapping from drug combinations to a RR can be turned into a binary classification problem where a prediction lower bound over a threshold (e.g. $1.1$) corresponds to a PIP, else to a safe drug combination. Therefore, classical classification performance metrics such as precision and recall are used here to evaluate the model trained by OptimNeuralTS. In addition to these two metrics, we also report the ratio of dangerous patterns used to generate the data that were found (Ratio $\mathcal{P}$), as well as the ratio of PIPs detected in $\mathcal{D}$ that are not in $\mathcal{D}_T$ (Ratio $\not\in\mathcal{D}_T$). These last two metrics aim to evaluate the generalization to PIPs unseen during training as the dangerous patterns $\mathcal{P}$ are not in $\mathcal{D}$ but are still known to have a high RR. All the results are reported for 25 repetitions of the experiments and every evaluation metric is computed at every 200 time steps of training. Furthermore, in order to quantify the benefits of using an ensemble model we compare the metrics of the ensemble approach to that of the latest trained model (i.e. the single model trained by OptimNeuralTS before the evaluation).

\subsubsection{Implementation details}
Since feature vectors are multi-hot vectors, the recommended drug combination ($\hat a_t$) is transformed into the most similar one in $\mathcal{D}$ ($a_t$) using the Hamming distance. Furthermore, unlike the original NeuralTS training routine that trains for a set amount of gradient steps and then returns the last parameters $\theta$ computed with the gradient step, our implementation keeps the parameters $\theta$ associated with the lowest loss on the training dataset as this maximizes likelihood~\cite{goodfellow2016deep}. As training is time consuming, the neural network is retrained every 10 steps and uses the Adam~\cite{kingma2014adam} optimizer due to its faster convergence than regular SGD. Finally, in order to simplify and accelerate computation, we rely on the same tricks as the original NeuralTS implementation~\cite{zhang2021neural}: we approximate the matrix $U$ by taking only its diagonal, remove the division by $m$ (see Eq.~\ref{eqn:std}) and only compute the L2 penalty on the current weights $\theta_t$.

\section{Results}
\label{sec:results}

\begin{figure}[t]
    \centering
    \begin{minipage}{\textwidth}
        \leftfigure{
            \includegraphics[width=0.49\textwidth]{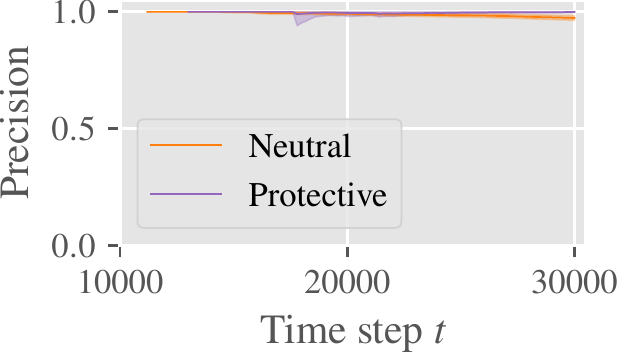}
        }
        \hspace{\fill}
        \rightfigure{
            \includegraphics[width=0.49\textwidth]{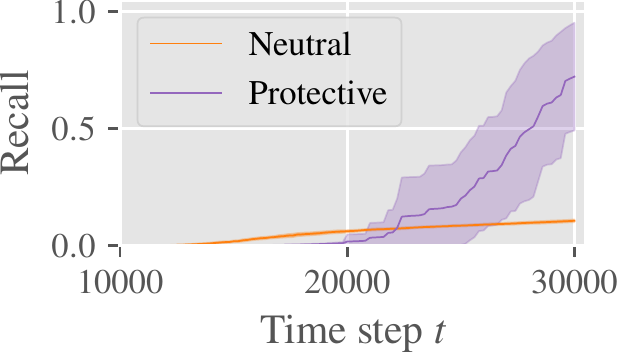}
        }

    \end{minipage}
    \leftcaption{Precision on the neutral and protective instances using the ensemble predictive model}\label{fig:precision_all}
    \rightcaption{Recall on the neutral and protective instances using the ensemble predictive model}\label{fig:recall_all}
\end{figure}

Figures~\ref{fig:precision_all} and~\ref{fig:recall_all} respectively display the precision and recall of the ensemble predictive model produced using the OptimNeuralTS training procedure over the growing time horizon. We observe a monotonic improvement of the recall over time in both settings, while managing to keep the precision excellent. The lower recall and slightly lower precision in the neutral instance is expected and is due to the possibility of the observed RR crossing the threshold because of the noise term $\xi_t$. Indeed, as time progresses, the agent must focus on drug combinations near the RR threshold, thus leading to slightly more false positives.

\begin{figure}[t]
    \centering
    \begin{minipage}{\textwidth}
        \leftfigure{
            \includegraphics[width=0.49\textwidth]{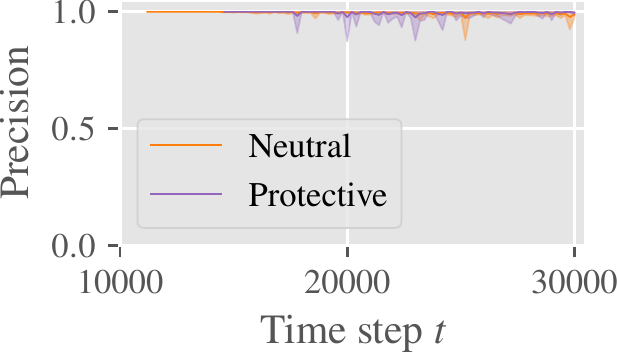}
        }
        \hspace{\fill}
        \rightfigure{
            \includegraphics[width=0.49\textwidth]{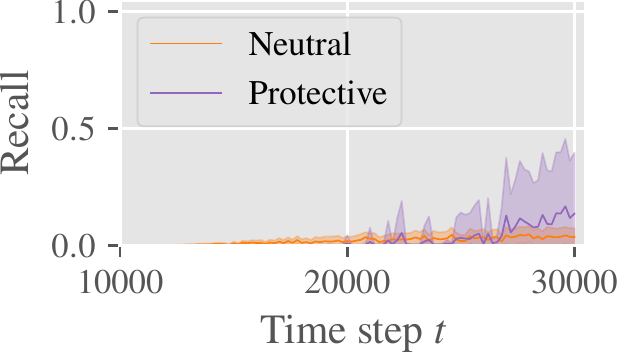}
        }

    \end{minipage}
    \leftcaption{Precision on the neutral and protective instances using the single latest model}\label{fig:precision}
    \rightcaption{Recall on the neutral and protective instances using the single latest model}\label{fig:recall}
\end{figure}

\subsection{Impact of ensemble} To highlight the benefits of using the history of generated models as an ensemble rather than simply relying on the most recent model, Figures~\ref{fig:precision} and~\ref{fig:recall} respectively show the precision and recall obtained at each time step by using only the most recent model instead of the ensemble. We first observe (Fig.~\ref{fig:recall}) that the most recent model is not consistently getting better at detecting PIPs on its own. Indeed, later models sometimes have worse recall than those of earlier iterations, suggesting that they did not retain knowledge acquired earlier. However, we observe (Fig.~\ref{fig:precision}) that high precision is maintained throughout the time steps, although with more noise compared with the ensemble (Fig.~\ref{fig:precision_all}). That is, every individual neural network trained by OptimNeuralTS has a low false positive rate. The ensemble leverages this fact efficiently by using a single vote to classify a combination as potentially harmful.
Moreover, we observe in general that the precision for single models fluctuates more for the protective instance than for the neutral instance. This behavior is most likely due to the wider range of RR of the protective dataset which results in drug combinations having similar components but very different RR. Even so, the precision remains high enough to have very little false positives in practice for both datasets when using ensembles. Indeed, the ensemble approach results in $819 \pm 47$ correctly detected PIPs for $20 \pm 9$ false positives on the neutral instance while it yields on the $1504 \pm 481$ correctly detected PIPs for $1 \pm 0$ false positives on the protective instance. The bigger fluctuation in the number of true positives for the protective instance can also be attributed to the bigger range of RR to cover.

\subsection{Generalization} 
\begin{figure}[t]
    \centering
    \begin{minipage}{\textwidth}
        \leftfigure{
            \includegraphics[width=0.49\textwidth]{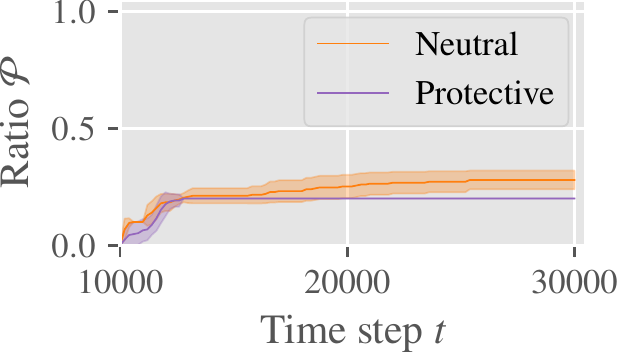}
        }
        \hspace{\fill}
        \rightfigure{
            \includegraphics[width=0.49\textwidth]{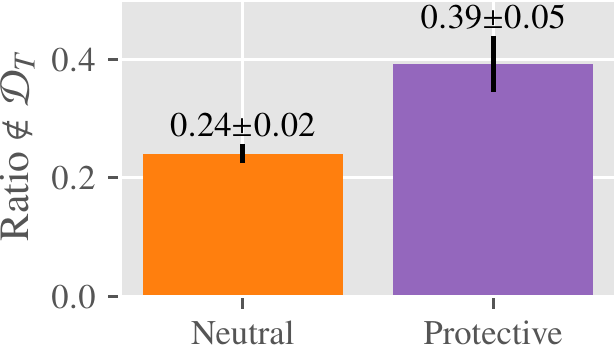}
        }
    \end{minipage}
    \leftcaption{Ratio $\mathcal{P}$ on the neutral and protective instances using the ensemble predictive model}\label{fig:ratio_p_all}
    \rightcaption{Ratio $\not\in\mathcal{D}_T$ on the neutral and protective instances using the ensemble predictive model}\label{fig:seen_pips}
\end{figure}

Figure~\ref{fig:ratio_p_all} shows the ratio of dangerous patterns $\mathcal{P}$ identified as PIPs by the resulting model. As previously mentioned, the dangerous patterns used to generate data are not in $\mathcal D$. However, although they can never be observed directly, the ensemble is capable of detecting dangerous patterns.  Furthermore, we observe that a good percentage of the detections made by the ensemble are not contained in the training data $\mathcal{D}_T$.

We observe (Fig.~\ref{fig:seen_pips}) that on average up to 39\% (on the protective instance) of detected PIPs were not even in $\mathcal{D}_T$ upon the completion of training. This is promising, as it indicates that the resulting ensemble can pick up on observed trends to predict unseen patterns. In practice, this would represent drug combinations that have never been prescribed together before, but whose combination could be dangerous, and so detecting them will prevent health risks for patients.

\section{Related Work}
\label{sec:relatedworks}

There exists prior work on data mining in the polypharmacy context. Methods have been proposed to detect new potentially inappropriate medications or model the association of polypharmacy to side effects using small datasets~\cite{thomas2020potentially, held2017polypharmacy}. General machine learning techniques have also been used to model polypharmacy and its side effects~\cite{zitnik2018modeling, masumshah2021neural,lakizadeh2022detection} by using complex drug data that are usually not contained in claims database. Therefore, efficiently learning from large datasets such as those considered in the current work require new approaches, hence motivating the proposed bandit angle.

Several contextual bandit strategies have been proposed previously to handle combinatorial problems with linear rewards~\cite{cesa2012combinatorial, combes2015combinatorial}. The linear reward assumption is common and allows for efficient computations of action recommendations. However, the tackled application requires the estimation of a (possibly) non-linear reward functions. Although there exists non-linear reward combinatorial bandit strategies~\cite{chen2013combinatorial}, they rely on oracles to recommend an action to play. Such oracles are typically designed for specific problems and unfortunately, our problem is not one of them. Alternatively, strategies to extract the top-$K$ best actions~\cite{rejwan2020top} do not assume linearity of the reward function and do not rely on an oracle. However, they require to preset the $K$-order magnitude of relevant actions, which is a priori unknown in the current application. Considering $K$ too low would result in missing PIPs, while setting it too high would result in false positives.

As our objective is not to maximize the cumulative rewards (Eq.~\ref{eqn:obj_bandit}), but rather explore drug combinations in order to detect potentially dangerous ones, the pure exploration setting would also be a natural formulation for the current application. Several combinatorial pure exploration bandit strategies have been proposed~\cite{chen2014combinatorial,du2021combinatorial, pmlr-v65-chen17a} previously. However, due to their combinatorial nature, they all exhibit a dependency on an oracle, which makes them then unusable in the tackled problem.
Pure exploration neural bandits have also been studied previously~\cite{zhu2021pure}. Although theoretically relevant, these methods bear important implementation challenges that prevent them from being used efficiently. The proposed OptimNeuralTS strategy is simpler to implement while still maintaining high precision and a capacity to generalize to unseen data.

Finally, thresholding bandits~\cite{pmlr-v48-locatelli16} is another relevant setting where the objective is to extract actions with a mean value estimate over a certain threshold. However, proposed approaches for this setting~\cite{pmlr-v48-locatelli16,mukherjee2017thresholding} only maintain mean estimates of every actions encountered during the game and could therefore not lead to a model that can predict the association measure for any new drug combination. Without such a model it becomes impossible to generalize to unseen actions as required by the tackled application.

\section{Conclusion}
\label{sec:conclusion}

This paper introduces the OptimNeuralTS approach combining NeuralTS~\cite{zhang2021neural} and differential evolution~\cite{storn1997differential} to data mine relevant data from very large unbalanced datasets. This method leverages the neural contextual bandit formulation to create an information-rich dataset on which to learn an ensemble predictive model. OptimNeuralTS is a general method for data mining that can be applied to any unlabelled dataset with a combinatorial structure. We conduct experiments using simulated datasets representing both protective and neutral settings. Results show that the predictive model learned with OptimNeuralTS is empirically capable of detecting PIPs with high precision. More importantly, the model is able to identify underlying dangerous patterns that are not observed directly in the data. These encouraging results suggest that OptimNeuralTS is a promising approach for guiding pharmaceutical research by recommending potentially dangerous drug combinations to investigate further and therefore contribute to safer prescriptions.

In future work, one could attempt to improve the sample efficiency using pure exploration neural bandits methods. Furthermore, while we do not take them into account, other important factors other than the presence of drugs (e.g. sex, age, medical conditions) contribute to whether a combination should be considered a PIP. Therefore, our simulation data can still be improved to portray a more complete setting.

\section{Acknowledgments}

This project is supported by the Canadian Institute of Health Research and the Natural Sciences and Engineering Research Council of Canada, grant number CPG-170621. Caroline Sirois receives a Junior 2 salary award from the Fonds de recherche du Qu\'ebec-Sant\'e. We would also like to thank CIFAR for the CCAI Chair funding.

\bibliographystyle{abbrv}
\bibliography{bibliography}

\end{document}